\documentclass[journal]{IEEEtran}

\usepackage{cite}

\ifCLASSINFOpdf
  \usepackage[pdftex]{graphicx}
  \usepackage{graphics}
  \DeclareGraphicsExtensions{.pdf,.jpeg,.png}
\else
  \usepackage[dvips]{graphicx}
  \usepackage{graphics}
  \DeclareGraphicsExtensions{.eps}
\fi

\usepackage{amsmath}
\usepackage{amssymb,amsthm}
\usepackage{algorithm}
\usepackage{algorithmic}

\usepackage{threeparttable}
\usepackage{array}
\usepackage{multirow}
\usepackage{colortbl,booktabs}

\ifCLASSOPTIONcompsoc
  \usepackage[caption=false,font=normalsize,labelfont=sf,textfont=sf]{subfig}
\else
  \usepackage[caption=false,font=footnotesize]{subfig}
\fi

\ifCLASSOPTIONcaptionsoff
  \usepackage[nomarkers]{endfloat}
  \let\MYoriglatexcaption\caption
  \renewcommand{\caption}[2][\relax]{\MYoriglatexcaption[#2]{#2}}
\fi

\usepackage{url}

\usepackage{pgfplots}
\pgfplotsset{compat=1.13}

\usepackage{afterpage}

\begin{document}

\title{On the Diversity of Realistic Image Synthesis}

\author{Zichen~Yang,
        Haifeng~Liu,~\IEEEmembership{Member,~IEEE}
        and~Deng~Cai,~\IEEEmembership{Member,~IEEE}
\thanks{Corresponding author: D. Cai.}%
\thanks{D. Cai is with the State Key Laboratory of CAD\&CG, College of Computer Science, Zhejiang University, Hangzhou, Zhejiang 310058, China(emails: microljy@zju.edu.cn, dengcai@gmail.com).}%
\thanks{Z. Yang and H. Liu are with College of Computer Science, Zhejiang University, China(emails: zichenyang.math@gmail.com, haifengliu@zju.edu.cn).}}

\ifCLASSOPTIONpeerreview
\markboth{IEEE Transactions on Image Processing}{Yang \MakeLowercase{\textit{et al.}}: On the Diversity of Realistic Image Synthesis}
\fi

\maketitle

\begin{abstract}
Many image processing tasks can be formulated as translating images between two image domains, such as colorization, super resolution and conditional image synthesis. 
In most of these tasks, an input image may correspond to multiple outputs. However, current existing approaches only show very minor diversity of the outputs. 
In this paper, we present a novel approach to synthesize diverse realistic images corresponding to a semantic layout. 
We introduce a diversity loss objective, which maximizes the distance between synthesized image pairs and links the input noise to the semantic segments in the synthesized images. 
Thus, our approach can not only produce diverse images, but also allow users to manipulate the output images by adjusting the noise manually. 
Experimental results show that images synthesized by our approach are significantly more diverse than that of the current existing works and equipping our diversity loss does not degrade the reality of the base networks.
\end{abstract}

%

\IEEEpeerreviewmaketitle

\section{Introduction}
\label{sec::introduction}
\IEEEPARstart{M}{any} problems in image processing and computer vision can be formulated as designing a map transforming an image from one domain to another domain, such as colorization\cite{zhang2016colorful,iizuka2016let}, super resolution\cite{timofte2014a+,dong2016image,kim2016accurate}, style transfer\cite{gatys2016image,luan2017deep} and conditional image synthesis\cite{mirza2014conditional,yan2016attribute2image,odena2016conditional,li2016combining,zhang2016stackgan}. 
Recently, Isola et al.\cite{isola2016image} tackle these problems by training a conditional generative adversarial network(GAN)\cite{Goodfellow2014Generative} with pairs of images from two domains.

We note that, most of these tasks have the nature that a single input may corresponds to mutiple solutions. 
For example, given a greyscale image containing a woman, there are various reasonable ways to fill the color of her hair and clothes. 
Therefore, the conditional GAN is fed not only an input image but also random noise, so that it can produce the corresponding images stochastically and moreover the produced images are diverse. 
However, Isola et al. observe that their models tend to simply ignore the noise, leading to very minor diversity of their output images\cite{isola2016image}. 
Thus, there are some following works\cite{deshpande2016learning,cao2017unsupervised,guadarrama2017pixcolor,chen2017photographic} trying to deal with this issue (Most of them focus on the colorization task. \cite{chen2017photographic} does the same task as ours). But, currently there is not a satisfactory solution.

\begin{figure}
\centering
\label{fig::intro}
\includegraphics{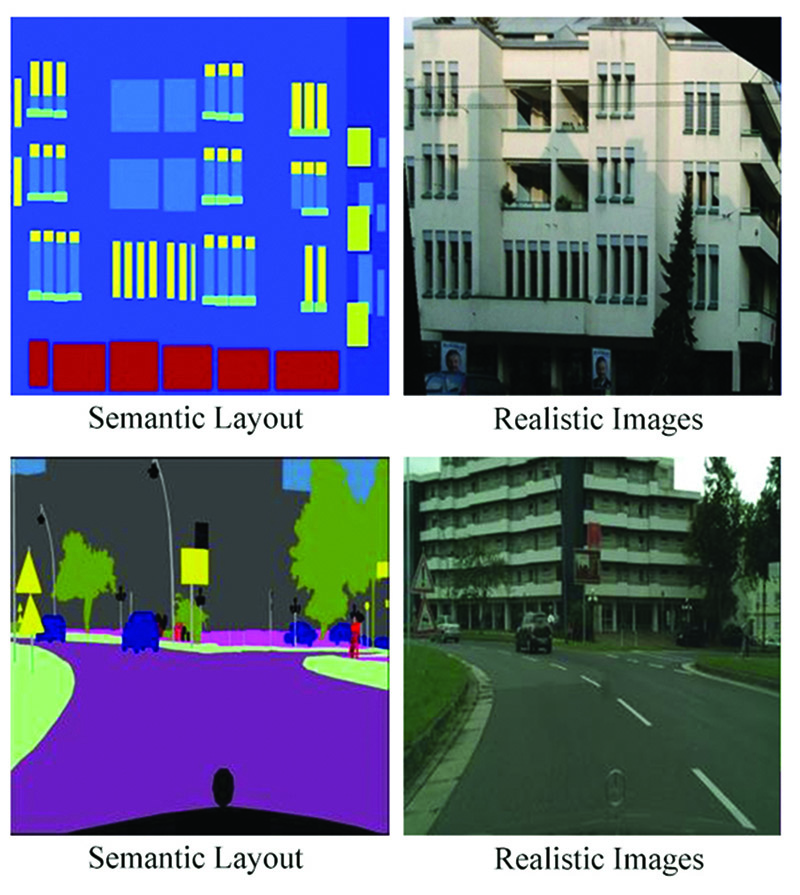}
\caption{These are two pairs of semantic layouts(each color represents a semantic class) and the corresponding realistic images from the CMP Facades dataset\cite{tylecek13} and the Cityscapes dataset\cite{Cordts2016Cityscapes}, respectively. Our task is to learn a map from the left domain to the right one.}
\end{figure}

In this paper, we give a novel approach to this issue focusing on a subfamily of the tasks, \textit{i.~e.,} transforming semantic segmentation layouts(images in the left column of Figure 1) to photo realistic images(images in the right column of Figure 1). 
The requirement of generative diversity naturally arise when dealing with these tasks.
For example, given a sketch of a building, it is better to synthesize multiple different designing proposal so that users can probably pick up a satisfying designing.
Moreover, one of the central problems from computer graphics also connects to these tasks.
It is suggested that directly synthesizing realistic images could act as a complement of current rendering engines, which avoids computationally intensive light transport simulation\cite{chen2017photographic}.

Our idea to generate diverse images is simple. 
Given a semantic layout $l$ and a random noise $n$, the task is to learn a map $f\colon L\times N\to I$, where $L$ is the space of semantic layouts, $(N, d)$ is the space of the random noise with a metric $d$ and $(I, \delta)$ is the space of realistic images with a metric $\delta$. 
Let $f_l$ be the restriction of $f$ to $\{l\}\times N$. 
To ensure that the outputs are diverse, it is equivalent to require that the distance of the outputs $\delta(f_l(l, n_1),f_l(l, n_2))$ is large for any inputs $n_1\neq n_2\in N$ and any $l\in L$.

In our approach, we slightly modify this idea to a conditional one, \textit{i.~e.,} $\delta(f_l(n_1),f_l(n_2))$ increases as the distance of noise $d(n_1,n_2)$ increases. 
This means that the input noise is linked to the synthesized images. 
Thus, we can not only generate diverse images, but also manipulate the synthesized images by adjusting the noise manually. 
In addition, our approach actually links each dimension of the noise to a semantic segment. 
Thus, slightly varying one dimension by keeping other constant can adjust the corresponding segment independently. 
This property could allow for applications where users can manipulate the image like tuning sliding knobs.
Moreover, due to this, we can generate a large number of diverge images, which is exponential to the number of semantic classes. 
Indeed, if a semantic layout contains $n$ classes and each dimension of the noise takes $k$ valid values. 
The number of images is the composition number $k^n$.

Another advantage of our approach is that this idea is realized by adding a diversity loss objective. 
Hence, this idea can easily be applied to most of the generative networks without much extra modification, just like an independent plug-in module. 
We shall test this property on two base models\cite{isola2016image,chen2017photographic}.

Finally, we evaluate both the reality and the diversity of the images synthesized by our approach using the datasets, the CMP Facades dataset\cite{tylecek13} and the Cityscapes dataset\cite{Cordts2016Cityscapes}.
Comparing to the base models\cite{isola2016image,chen2017photographic}, the experimental results indicate that our approach can produce more diverse images meanwhile preserving the same level of reality.

\section{Related Works}
The basic model of this paper, GAN\cite{Goodfellow2014Generative} is proposed to generate realistic images. 
The original GAN has many drawbacks, such as mode collapse, parameter sensitivity and poor quality. 
Thus, many advanced techniques\cite{denton2015deep,Radford2015Unsupervised,arjovsky2017wasserstein,chen2016infogan} has been applied to improve the process of image synthesis by GANs. 
For a comprehensive tutorial, see \cite{goodfellow2016nips}.

Isola et al. are the first ones that consider image-to-image translation problems in a unified view and present a conditional GAN\cite{mirza2014conditional} based framework, Pix2pix, to tackle these problems\cite{isola2016image}. 
They show that U-Net\cite{ronneberger2015u} is a better choice for the generator than DCGAN\cite{Radford2015Unsupervised} and apply an adversarial discriminator to learn various structured losses for image modeling. 
Since paired data is not always available, three similar works\cite{zhu2017unpaired,yi2017dualgan,kim2017learning} give approaches to do image-to-image translation with only unpaired data.

However, all of these models cannot generate diverse outputs corresponding to an input, which contradicts the multiple-solution nature of image-to-image translation tasks. 
Hence, much literature has been devoted to address this issue. 
Most of the existing works focus on the task of image colorization. 
For example, \cite{deshpande2016learning} build a multi-modal distribution between greyscale image and the color field embeddings, which results in diverse colorization. 
\cite{cao2017unsupervised} develop a fully convolutional generator with multi-layer noise to enhance diversity. 
\cite{guadarrama2017pixcolor} train a conditional pixelCNN to produce more diverse and plausible colorizations.

Another work\cite{chen2017photographic} by Chen el al. does the same task as ours, \textit{i.~e.,} translating semantic layouts to realistic images. 
They use a deterministic network instead of a GAN. 
Their architecture produces $n$ images in the final layer and they only minimizes the distance between the best one and the ground truth. 
They claim that this encourages the network to spread its bets and produce diverse results. 
But in their experiment, the multiplicity $n$ is set to only $9$. 
That is to say, if the users want to produce $n_0$ different images with $n_0$ much more than $9$, the users have to retrain a new network that outputs $n_0$ images in the final layer. 
We observe that this will cause very heavy visual memory overhead, which shall be discussed in detail later.

Different from this work, our model applies the stochastic model to avoid heavy visual memory overhead and further explicitly emphasizes that the distance of two synthesized images should be large conditioning on the distance of the noise. 
Later, our experimental results will show that this method causes better diversity. 
In addition, our model links the random noise to the distance. 
Thus, the outputs can be manipulated by users.

\section{Method}


Since our approach has to compute the distance between noise and the output images, we assume that the space of random noise and realistic images are metric spaces. 
The complete list of notations is listed in Table 1.
\begin{table}
\centering
\caption{Notation frequently used through this paper} 
\label{tbl::notation} 
\begin{tabular}{|c|c|c|}
\hline  
\textbf{Notation} & \textbf{Meaning} & \textbf{Remarks} \\
\hline
\multicolumn{3}{|c|}{Spaces and sets} \\
\hline
$L$   & the space of semantic layouts    &  \\
$N$   & the space of random noise        & metric $d$ \\
$I$   & the space of realistic images    & metric $\delta$ \\
$C$   & the set of semantic classes       &  \\
\hline
\multicolumn{3}{|c|}{Elements} \\
\hline
$(l_k, i_k)\in L\times I$ & the $k$-th layout-image pair & \\
$c\in C$ & a semantic class &  \\
\hline  
\end{tabular}  
\end{table}

Given the dataset $\{(l_k, i_k)\}_{k=1}^{m}$, our task is to learn a map $f\colon L\times N\to I$ such that fixing the layout $l_k$, each map $f_{l_k}=f(l_k,\cdot)$ can produce diverse realistic images.

\subsection{Pix2pix}
The method of Pix2pix to learn this map is to train a conditional GAN that consists of a generator $G\colon L\times N\to I$ and a discriminator $D\colon L\times I\to\mathbb{R}$. 
The generator $G$ that plays the role of the map $f$ mentioned previously, is trained to produce images that cannot be distinguished from realistic images by the discriminator $D$. 
On the other hand, the discriminator $D$ is adversarially trained to distinguish the images synthesized by $G$ from realistic images. 
After the training process, if the discriminator fails to distinguish the images synthesized by $G$, $G$ can approximate the desired map $f$.
In practice, both $G$ and $D$ are parametrized deep neural networks.
The parameters are updated using gradient based optimization algorithms to optimize some specific objectives.

In order to measure the reality of images, $D$ should output high confidence when fed realistic images and output low confidence when fed images produced by $G$. 
Thus, the objective of $D$ is to maximize the loss function as follows.
\begin{equation}
\begin{split}
\mathcal{L}_D=&\mathbb{E}_{l,i\sim p_{L, I}}\log[D(l,i)]+\\
&\mathbb{E}_{l\sim p_{L};n\sim P_{N}}\log[1-D(l,G(l,n))],
\end{split}
\end{equation}
where $P_{L,I}, P_{L}$ and $P_{N}$ are the joint distribution of $I$ and $L$, the distribution of $L$ and $N$, respectively. 
We should note that all of these distributions are estimated by the given data empirically throughout this paper.

Similarly, the objective of the generator $G$ is to minimize the loss function as follows.
\begin{equation}
\begin{split}
\mathcal{L}_G=&~\mathbb{E}_{l\sim p_{L};n\sim p_{N}}\log[1-D(l,G(l,n))]+\\
&~\alpha\mathbb{E}_{l,i\sim p_{L,I};n\sim P_{N}}\lVert G(l,n)-i\rVert_1,
\end{split}
\end{equation}
where $\alpha$ is a hyperparameter and $P_{L, I}$ is the joint distribution of $L$ and $I$.
Minimizing the first term forces $G$ to generate more realistic images $D$ cannot distinguish. 
Minimizing the second term ensures that the output of $G$ is close to the ground truth in $L_1$ sense. 
We should note that Pix2pix uses $L_1$ loss, since some previous works\cite{ledig2016photo,larsen2015autoencoding} observe that $L_2$ loss usually leads to blurry images while $L_1$ loss produces more realistic images.

Hence, the final objective is
\begin{equation}
\mathop{\arg\max}_{D}\mathcal{L}_D+\mathop{\arg\min}_{G}\mathcal{L}_G.
\end{equation}
This multi-objective optimization problem is solved by optimizing two objectives using Adam optimizer\cite{Kingma2014Adam}, alternatively.

The main drawback of Pix2pix is that the training process tends to ignore the noise $n$ and the dropout randomness\cite{srivastava2014dropout}, resulting in very little stochaticity in the synthesized images. 
To avoid ignoring the noise, the simplest method is to add a penalty for the ignorance.
That is to say, we minimize a diversity loss function in addition to the current objective, encouraging the generator to pay attention to the noise and synthesize diverse results.

\subsection{Cascaded Refinement Network}
To avoid the drawback of Pix2pix, the Cascaded Refinement Network(CRN)\cite{chen2017photographic} applies a deterministic network instead to generate multiple images at one time, and forces the images to be diverse.
The main points of the CRN are stated as follows.

The CRN is a cascade of differentiable modules, each module containing several convolutional layer, layer normalization layers\cite{ba2016layer} and leaky ReLU layers\cite{maas2013rectifier}.
The first module $M_0$ transforms the downsampled semantic layout of size $w_0\times h_0$ to the outputs of size $w_1\times h_1$, where $w_1=2w_0$ and $h_1=2h_0$. 
For each $i\geq1$, the $i$-th layer module $M_i$ receives the semantic layout of size $w_i\times h_i$ and the outputs of $M_{i-1}$, and outputs feature maps of size $w_{i+1}\times h_{i+1}$, where $w_{i+1}=2w_{i}$ and $h_{i+1}=2h_{i}$. 
If we aim to generate realistic images of size $256\times 512$, One could set $w_0=4$ and $h_0=8$, and set the number of output channels of $M_6$ to $3$. 
Then it outputs an image of size $256\times 512$ with RGB color channels.

To make sure that the outputs are realistic, the CRN adopts the content loss proposed by \cite{gatys2016image}. 
In detail, let $\Phi_k$ be the $k$-th layer representation of a pretrained VGG-19 model\cite{simonyan2014very}. 
The content loss of CRN is
$$\mathcal{L}_{CRN}=\mathbb{E}_{l,i\sim p_{L, I}}\sum_{k}\lambda_k\lVert \Phi_k(i)-\Phi_k(G(l))\rVert_1,$$
where $G(\cdot)$ denotes the cascaded generator and each $\lambda_k$ is the hyperparameter to balance the contribution of each layer $k$ to the loss. 
Since each layer of VGG model extracts a certain level of abstraction of images, this loss can guide the network to concern both fine-grained details and more global part arrangement\cite{chen2017photographic}.

The current version of the CRN cannot produce diversity, since it is deterministic.
To overcome this, the number of output channels of the $M_6$ is modified from $3$ to $3n$ so that the CRN produces $n$ different realistic images of RGB channels.
Moreover, the content loss is also modified, which only penalizes the image that is closest to the ground truth.
The authors claim that this encourages the CRN to spread the generated images and produces diversity.

However, there is one disadvantage of the CRN, the visual memory cost. 
Indeed, even when the users set $n=9$, the training process takes up more than $6$ gigabytes of visual memory to generate $256\times512$ images. 
Now suppose that one has to produce $n_0$ different images with $n_0>>n$, the final layer of the network should be reset to $n_0$. 
Note that the final layer costs the major part of the visual memory because the size of outputs in the final layer are quite large, \textit{i.~e.,} $256\times512$.
Hence, even if we need a moderately larger diversity of images, say $n_0=64$, it is very difficult to put this huge network into GPUs.

Our approach can easily overcome the memory risk, since it only synthesizes $2$ images at one time, one generated with noise and the other without noise. 
Our diversity loss further explicitly emphasizes that the distance of the two images should be large conditioning on the norm of the noise. 
Thus, this encourages the image to vary diversely when one adjusts the input noise. 
The detailed idea will be illustrated next.

\subsection{Diversity Loss}
\label{sec::diversityloss}
Suppose that the generative base network is $G$($G$ of the Pix2pix or the CRN). 
Given a semantic layout $l$, assume that $d(n_1,n_2)=\lvert n_2-n_1\rvert$ and $I_l$ is the subspace of $I$ consisting of the outputs corresponding to $l$. 
As illustrated in Section \ref{sec::introduction}, our basic idea is that $\delta(G(l,n_1),G(l,n_2))$ should increase as $d(n_1,n_2)$ increases for $n_1\neq n_2\in N$.
In our approach, $n_1$ is always set to $0$ and we only let $n_2$ vary as variable. 
Thus, for simplicity, we change the notation $n_2$ to $n$.

In a single feedforward, we let $G$ produces two synthesized images using $(l, 0)$ and $(l, n)$. 
To encourage a large distance $\delta(G(l, 0),G(l, n))$, we minimize the negative distance between $G(l, 0)$ and $G(l, n)$ with respect to $d(n, 0)=\lvert n\rvert$, defined as
\begin{equation}
\label{eqn::prediversityloss}
\mathcal{L}_{Div}''=\mathbb{E}_{l\sim p_L;n\sim p_N}-\lvert n\rvert\lVert G(l, 0)-G(l, n)\rVert_1.
\end{equation}
To understand the effect of the factor $\lvert n\rvert$, consider the combined objective of $G$
\begin{equation}
\mathop{\arg\min}_{G}\mathcal{L}_B+\mathcal{L}_{Div}'',
\end{equation}
where $\mathcal{L}_B$ is the generative loss of the Pix2pix or the CRN, \textit{i.~e.,} $\mathcal{L}_G$ or $\mathcal{L}_{CRN}$.
One notices that if $\lvert n\rvert$ increases, the optimizer will be encouraged to minimize the second term. 
Thus, it will increase $\lVert G(l, 0)-G(l, n)\rVert_1$. 
This coincides with our requirement that $\delta(G(l, 0), G(l, n))$ should increase conditionally as $d(0, n)$ increases.

Moreover, we can squeeze out one more from this idea. 
Let $n$ be a $\lvert C\rvert$-dimensional noise vector drawn from $[-1,1]^{\lvert C\rvert}$ uniformly. 
To avoid confusing, remember that we use superscript $n^c$ to denote the $c$-th dimension of $n$. 
Let $\lVert\cdot\rVert_1^{c}$ be the segmentwise $L_1$ loss with respect to $c\in C$, that is to say, it only cumulates the $L_1$ difference of the pixels with the semantic label $c$. 
Then, the segmentwise diversity loss is defined as
\begin{equation}
\mathcal{L}_{Div}'=\mathbb{E}_{l\sim p_{L};n\sim p_N}-\sum_{c\in C}\lvert n^c\rvert\lVert G(l, 0)-G(l, n)\rVert_1^c.
\end{equation}
The only difference from Equation (\ref{eqn::prediversityloss}) is that we link each $n^c$ to one segment in the synthesized image. 
Thus, if we increase $\lvert n^c\rvert$, only the $c$-th segment in the synthesized images will be influenced. 
This encourages that the random noise $n^c$ is linked to the segment $c$. 
Hence, in testing time, we can obtain our desired image by adjusting each $n^c$ manually.

Due to this, we can generate a large number of diverge images, which is approximately exponential to the number of semantic classes. 
Assume that the semantic layout contains $|C|$ classes and each dimension of the noise $n^c$ can take $k_c$ valid values with a uniform lower bound $k\leq k_c$. 
Then, the number of images is the number of compositions $\prod_{c\in C}k_c\geq k^{|C|}$.

\subsection{Consistency Constraint}
There is another modification that we shall add to the current objective to make sure that the output is consistent with the semantic layout $l$. 
Note that we do not give a bound to $\mathcal{L}_{Div}'$. 
So, the $G(l, n)$ can be quite different from the ground truth. 
This issue can be addressed by adding a constraint.

We set an upper bound for the distortion of each segment, \textit{i.~e.,} change $\mathcal{L}_{Div}'$ to $\mathcal{L}_{Div}$, which is defined as
\begin{equation}
\begin{split}
\mathcal{L}_{Div}=&\mathbb{E}_{l\sim p_L;n\sim p_N}\\
&\sum_{c\in C}\lvert n^c\rvert\max(0, \lambda_c-\lVert G(l, 0)-G(l, n)\rVert_1^c),
\end{split}
\end{equation}
where $\lambda_c$ is the upper bound for the distortion of the segment $c$ and the negative sign $\mathcal{L}_{Div}'$ is moved inside the maximum.

Hence, the final objective of our approach is formulated as
\begin{equation}
\mathop{\arg\min}_{G}(\mathcal{L}_B+\beta\mathcal{L}_{Div}),
\end{equation}
where $\beta$ is the hyperparameter. 
To optimize the base networks with the diversity loss, one can refer to the algorithm descrition in the appendix.

\section{Experiments}
In this section, we conduct experiments to evaluate our approach on two datasets\cite{tylecek13,Cordts2016Cityscapes}, comparing to two base networks\cite{isola2016image,chen2017photographic}. 
Typically in the study of image synthesis, one is interested in both the diversity and the reality of the synthesized images. 
Thus, our experiments consist of two parts.

First, as we state in Section \ref{sec::diversityloss}, we show that our approach can link each dimension of the input noise to one segment of the synthesized images. 
Thus, the generative network can produce a diverse set of images whose cardinality is exponential to the number of semantic classes(shown in Section \ref{sec::diversity}).
Hence, the base networks together with our diversity loss can synthesize much more diverse images(shown in Section \ref{sec::diversity}).

Second, note that regardless of the reality of the images, one can achieve diversity easily by outputing random images.
Thus, we conduct experiments not only on the diversity of the generated images, but also the reality of the images. 
Indeed, we also show that this diversity loss does not degrade the reality of the original networks(shown in Section \ref{sec::reality}).

\subsection{General Settings}
We slightly modify the network settings and the training settings of two base networks.
First, to equip the base networks with the diversity loss, we do the minimal modification that is to add one more input channel to receive the input noise
The noise channel has the same size as the input image and we assign the value $n^c$ to one pixel in the noise channel, if its corresponding pixel belongs the semantic class $c$.
The other setting of the networks follows from the original papers\cite{isola2016image,chen2017photographic}. 
Second, there are two hyperparameters in our final objective. 
We set $\beta=0.1$ for the Pix2pix and $\beta=10$ for the CRN. 
For simplicity, we set $\lambda_c=0.3$ for all $c$ in two tasks. 
Finally, the total number of epochs to training the CRN is originally $200$, we set it to $100$.

There are two datasets we shall use, including the CMP Facades dataset\cite{tylecek13} and the Cityscapes datasets\cite{Cordts2016Cityscapes}. 
The CMP Facades dataset used in our experiments is the one preprocessed by Isola et al.\cite{isola2016image}. 
It contains $606$ pairs of semantic layouts and real images of facades, where $400$ for training, $100$ for validation and $106$ for testing, and all of the training pairs are cropped to the size $256\times256$. 
There are in total $12$ semantic classes, \textit{i.~e.,} $\lvert C\rvert=12$. 
We run the same experiments on a subset of the Cityscapes dataset.
It consists of $2975$ training examples and $500$ testing examples, where each one is rescaled to $256\times512$.
But for better presentation of cityscapes images, we halve the width of the output images.
The total number of semantic classes of the Cityscapes dataset is $19$.

To train the Pix2pix and the Pix2pix with the diversity loss, data augmentation is performed. 
Specifically, we resize every pair to be of size $286\times286$ using bicubic interpolation and extract a random patch of size $256\times256$. 
In addition, a random flip operation is performed on the extracted patches.

\begin{figure*}[t!]
\centering
\label{fig::facades}
\includegraphics[width=0.93\textwidth]{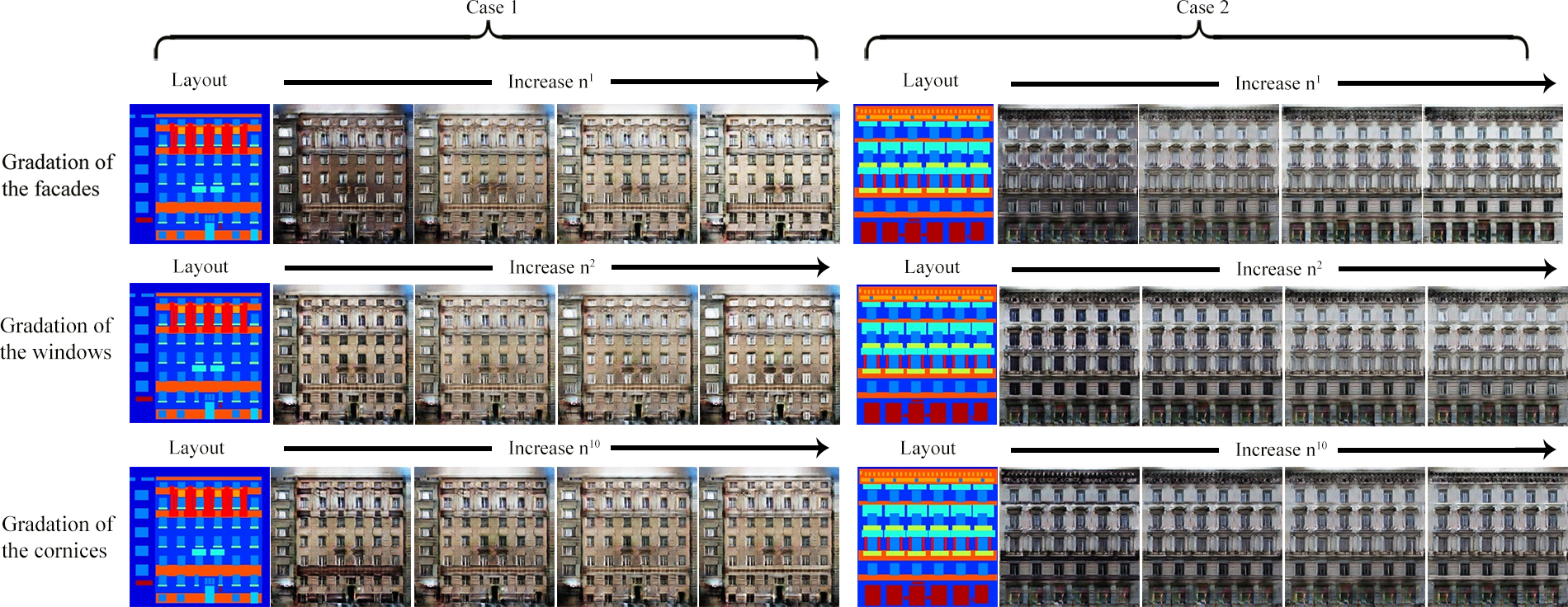}
\caption{\textbf{Gradation of segments of the CMP Facades dataset}: There are in total $12$ semantic classes of the CMP Facades dataset. 
We only show the results of the $1$st, the $2$nd and the $10$th classes, which represent facades, windows and cornices, respectively. 
The first row of images shows that the illuminance of the facades gradually increases as the linked random noise $n^1$ increases. 
Meanwhile, the other segments remain unchanged. 
Similarly, one can observe similar gradation of windows in the second row when increasing $n^2$, and gradation of cornices in the third row when increasing $n^{10}$.}
\end{figure*}

\begin{figure*}[h!]
\centering
\label{fig::cityscapes}
\includegraphics[width=0.93\textwidth]{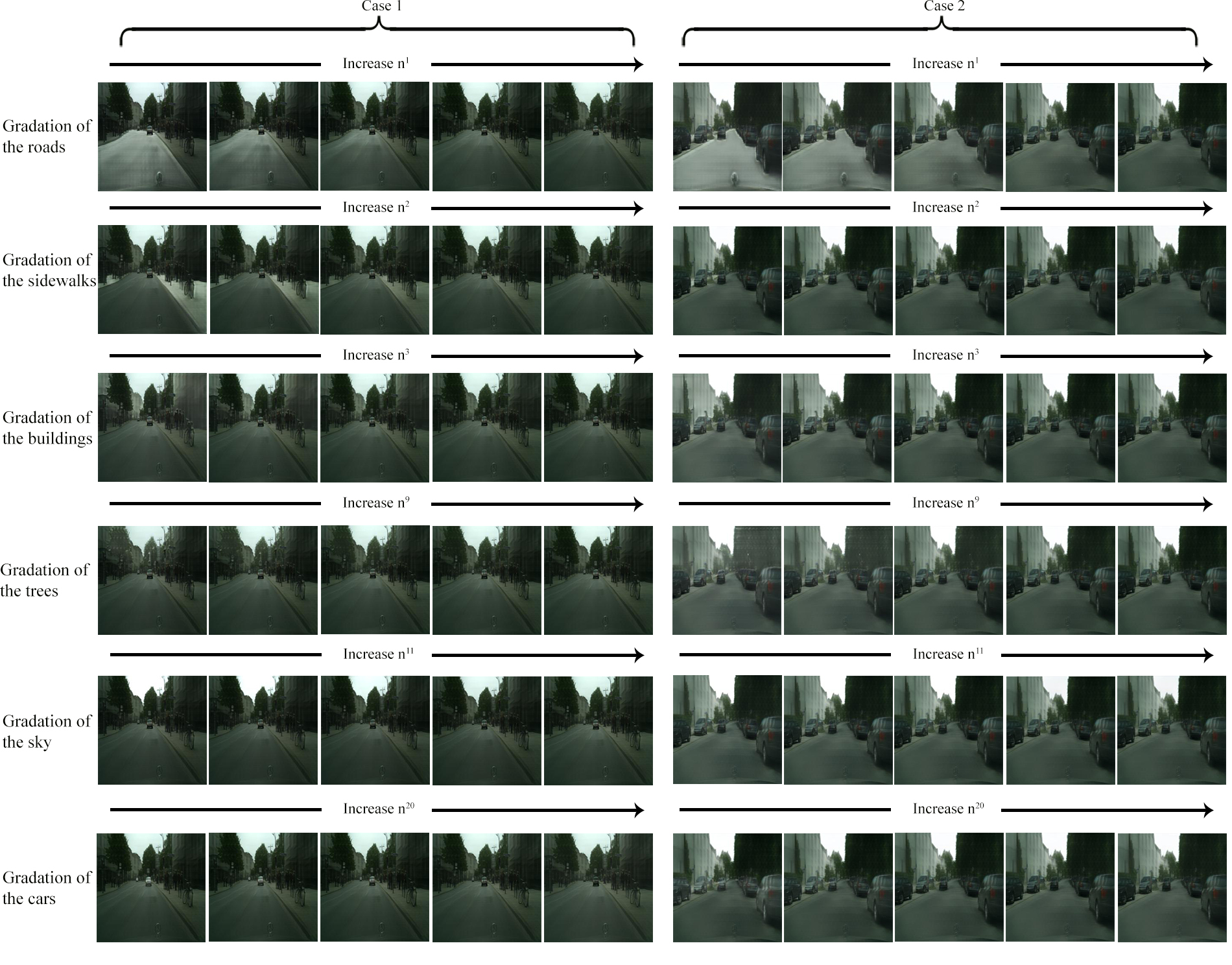}
\caption{\textbf{Gradation of segments of the Cityscapes dataset}: It is the experiments  on the Cityscapes dataset, similar to that in Figure 2. 
We only show the results of six classes, which take up most of the pixels, including the roads, the sidewalks, the buildings, the trees, the sky and the cars. 
One can observe similar gradation of segments.}
\end{figure*}

\subsection{Evaluating Diversity}
We are going to evaluate the diversity of our approach in two steps. 
First, we show that our approach can link each dimension of the noise to one segment, which can produce significantly more images than the base models. 
Second, we present the qualitative results of our approach.

\label{sec::diversity}
\noindent\textbf{Noise Linking}:
Our approach benefits from the linking between noise and the segments. 
Thus, one can produce a diverse set of images with cardinality being exponential to the number of semantic classes.
To support this claim, we slightly tune each dimension of the input noise while fixing the input semantic layout, and track the varying of the corresponding segments of the generated images.
Since usually some major segments will take up most of the space in one image, we only present the varying of these major segments for a better visualization.
For the CMP Facades dataset, we only present the results related to three semantic classes, including the facades, the windows and the cornices.
For the Cityscapes dataset, we present the result of the road, the sidewalk, the car, the building, the vegetation and the sky.

We present the results on the CMP Facades dataset in Figure 2. 
We collect two images in the test set for the varying of each dimension of the noise.
In Figure 2, from the left hand side to the right in each row, we gradually increase the value of one dimension of the noise. 
In the first row, one can observe that the illuminance of the facades gradually increases as $n^1$ increases from $-1$ to $1$, while the other segments are nearly unchanged.
Thus, $n^1$ can be explained as the relative illuminance of the facades. 
One can also find similar effects on the windows and the cornices, in the second row and the third row, respectively.
This phenomenon meets with our former claim that each dimension of the noise controls one segment. 

Similarly, the results on the Cityscapes dataset are presented in Figure 3.
We can also notice the same kind of gradual varying in one segment corresponding to the varying dimension of the noise.
At the same time, other segments are nearly fixed.

\begin{figure}[t]
\centering
\label{fig::crn_produce}
\includegraphics{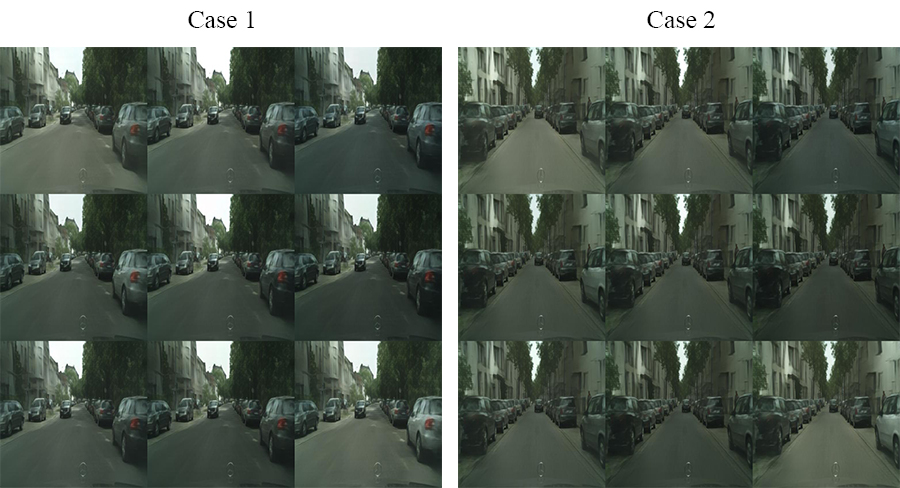}
\caption{\textbf{Two typical cases of the outputs of the CRN}: We use the pretrained CRN, which produces $9$ images in a single feedforward. 
We organize that the outputs in a $3\times 3$ array and two cases are presented.}
\end{figure}

\vspace{0.5cm}\noindent\textbf{Comparison}: 
We should compare to the results of the Pix2pix and the CRN, both without the diversity loss.
As written in the paper of the Pix2pix, the authors find very little stochasticity of the output images as different noise are fed\cite{isola2016image}.
So, the original Pix2pix cannot produce diverse sets of images.

As for the CRN, we download the pretrained CRN, which has $9$ images outputs. 
We organize the $9$ output images in the form in Figure 4. 
It turns out that the CRN actually learns $9$ image patterns. 
For example, the upper left one always produces an image with a relatively lighter road and lighter cars in both two cases while the upper right one with a darker road. 
Thus, the pretrained CRN has the diversity $9$, \textit{i.~e.,} it only outputs images from $9$ patterns.
If the users further require more images patterns, they have to retrain a new CRN with more output channels.

However, since the images in the output layer is very large, for instance, $256\times512$ or $512\times1024$, adding more output channels will lead to the risk of visual card memory.
Instead, using our diversity loss, the diversity is produced via random noise.
Thus, the output channel can be set to $1$, which avoids severe visual memory overhead.
We present the qualitative result of the diversity of our methods next.

\vspace{0.5cm}\noindent\textbf{Qualitative Study}:
Since every dimension of the random noise controls the illuminance of one specific segment, by adjusting the noise manually, one can manipulate the output images.
Thus, the users can synthesize many different images by our approach. 
We qualitatively prove this claim by showing diverse sets of generated images.

The base network of facades generation and cityscapes generation are the generator of the Pix2pix and the CRN, respectively.
We present the generated images in Figure 6.
One notices that our method can lead to not only many variations of the color of the facades and the windows, but also some structural variations.
For more results, we present them in the appendix.

On the other hand, we observe that, although the Cityscapes dataset has more semantic classes than the CMP Facades dataset, the generated cityscape images are not more diverse than the facades images.
We guess that this is due to the training of the base model, the CRN, penalizes not only the $L_1$ loss but also the content loss\cite{chen2017photographic}.
This strict penalty makes the generated images are more closer to the ground truth.

\begin{figure}[t]
\centering
\label{fig::compare_seg}
\includegraphics{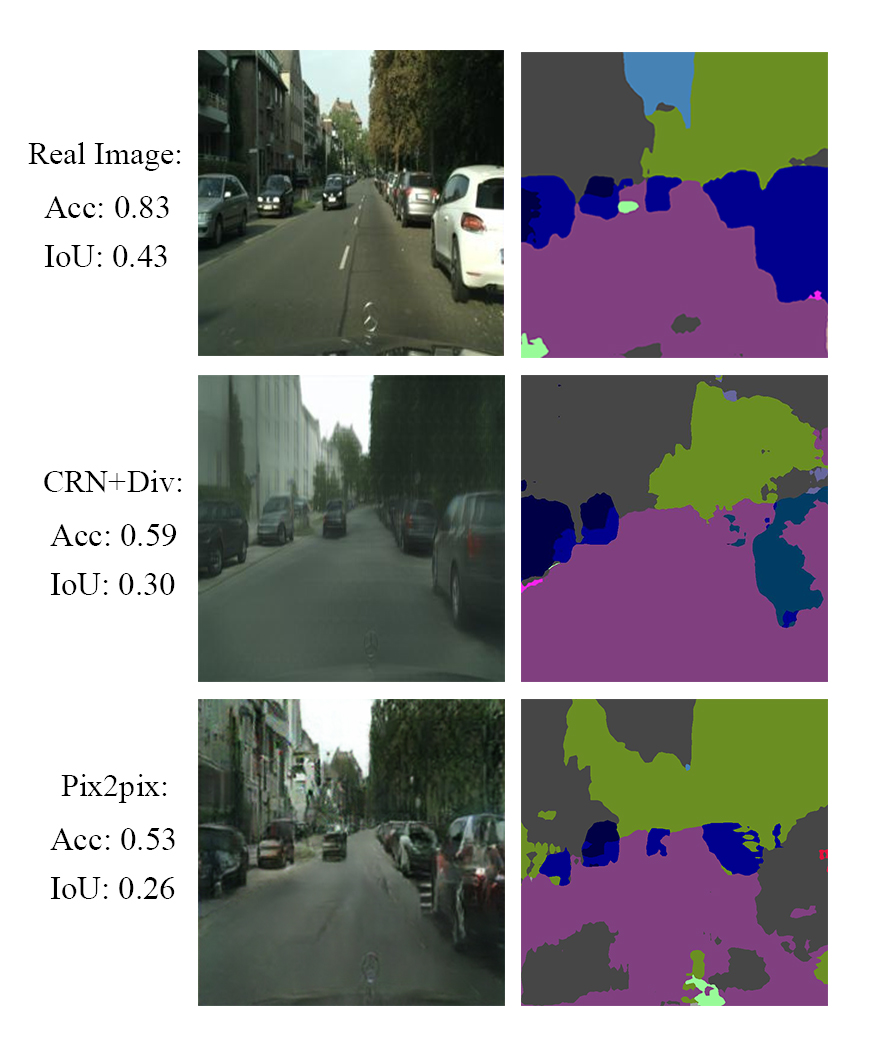}
\caption{\textbf{Examples of Semantic Segmentation Test}: We use a pretrained PSPNet\cite{zhao2017pspnet} to do semantic segmentation on the synthesized images and the real image. The second one uses the CRN base network with diversity loss. 
We calculate the Accuracy(Acc) and IoU of the predicted layouts. 
It turns out that, if the synthesized images are more realistic, the computed metric will be higher and the predicted layout will be more similar to the ground truth.}
\end{figure}

\begin{table}[!t]
\caption{Accuracy and IoU metric on Facades and Cityscapes}
\label{tbl::segmentation_test}
\begin{center}
\begin{tabular}{|c|c|c|c|c|}
 \hline
 \multirow{3}{*}{Model} & \multicolumn{4}{c|}{Datasets}\\
 \cline{2-5}
  & \multicolumn{2}{c|}{Facades} &
 \multicolumn{2}{c|}{Cityscapes} \\
 \cline{2-5}
   & Accuracy & IoU & Accuracy & IoU \\
 \hline
 Pix2pix & 0.464 & 0.103 & 0.619 & 0.315 \\
 Pix2pix+Diversity & 0.484 & 0.111 & N/A & N/A \\
 CRN & N/A & N/A & 0.721 & 0.344 \\
 CRN+Diversity & N/A & N/A & 0.704 & 0.361 \\
 \hline
\end{tabular}
\begin{tablenotes}
\footnotesize
\item 1. ``+Diversity'' means the base network equipped with the diversity loss. 
Since CRN is so large that it may overfit the CMP Facades dataset, we only run the CRN on the Cityscapes dataset.
\end{tablenotes}
\end{center}
\end{table}

\begin{figure*}[h!]
\centering
\label{fig::generation}
\includegraphics{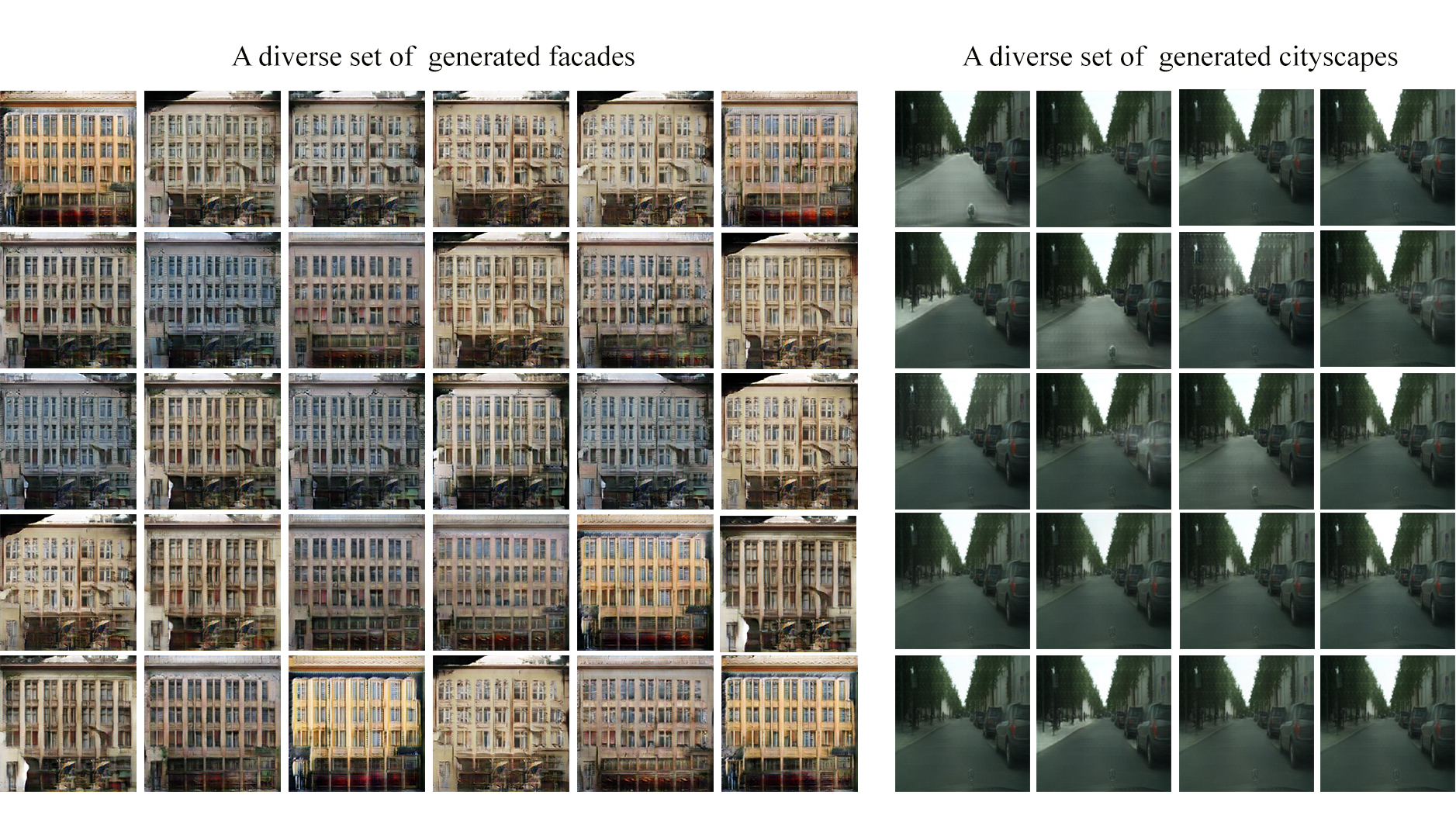}
\caption{\textbf{Diverse sets of facades and cityscapes generated by our approach}: The base network for the facades generation is the generator of the Pix2pix while the one for the cityscapes is the CRN. }
\end{figure*}

\subsection{Evaluating Reality}
\label{sec::reality}
\noindent\textbf{Quantitative Study}: 
Currently, there is not a systematic approach to test whether a given image is realistic or not or to evaluate the reality of a given image. 
But, thanks to the nature of our tasks, there is an empirical way to evaluate the reality of the synthesized images, since we have the ground truth semantic layout.
Indeed, one can pretrain a semantic segmentation network and feed it the synthesized images to receive a predicted semantic layout.
Intuitively, if the predicted layout is close to the ground truth layout, the synthesized images are realistic.

We calculate two common evaluation metric for semantic segmentation to measure the closeness of the predicted layout and the ground truth.
\begin{enumerate}
\item \textbf{Accuracy}: the proportion of the number of correctly predicted pixels to the number of the total pixels in one image.
\item \textbf{Intersection over Union(IoU)}: the number of the true positive pixels over the sum of the numbers of the true positive pixels, the false positive pixels and the false negative pixels.
\end{enumerate}

Actually, we observe that if the generated image is more realistic, the predicted layout will be probably similar to the ground truth semantic layout and the two metric will be higher.
Figure 5 gives an example that supports with our observation.

In our experiments, we use the SegNet\cite{badrinarayanan2015segnet} for the Facades dataset and the PSPNet\cite{zhao2017pspnet} for the Cityscapes dataset as the pretrained segmentation networks. 
We randomly select semantic layouts from two datasets, use them as inputs to generate realistic images, and finally produce the corresponding predicted layouts by the segmentation networks. 
Since both CRN and our approach can produce multiple images, for each semantic layout input, we take $9$ realistic images from the CRN and $32$ images from our approach.

The experimental results are shown in the Table \ref{tbl::segmentation_test}. 
It turns out that the networks equipped with the diversity loss objective have the same level of accuracy and IoU than the original networks. 
Thus, equipping our diversity loss objective does no harm to the reality of the base networks.
Thus, one can treat the diversity loss as an independent plugin module to enhance the generative diversity of many networks.

\vspace{0.5cm}\noindent\textbf{Qualitative Study}: 
Similar to evaluating diversity, we also conduct qualitative experiments on the reality.
We compare the images synthesized by our approach(base networks with the diversity loss) with the images synthesized by the base networks, the Pix2pix and the CRN.
The results are shown in the Figure 7.
From the presented comparison, one sees that equipping with the diversity loss will not degrade the reality of the base models.

\begin{figure}[!htb]
\centering
\label{fig::real}
\includegraphics{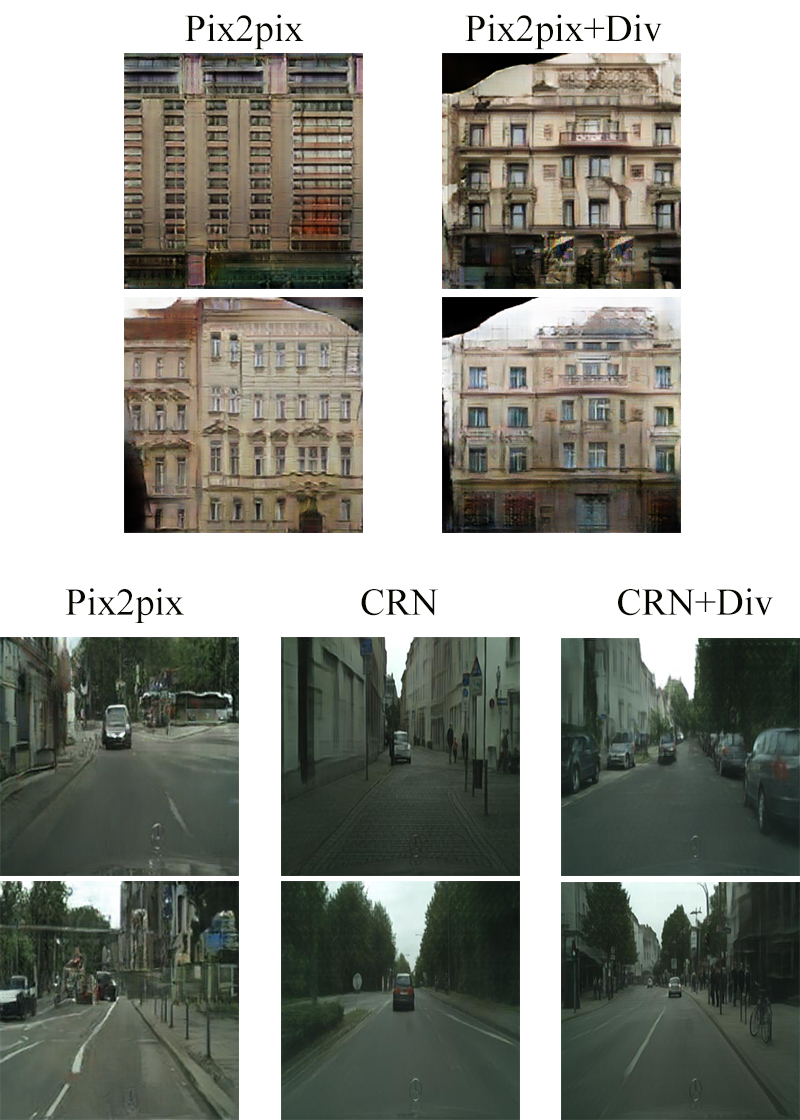}
\caption{\textbf{Comparing reality of the Pix2pix, the CRN and our methods}: The base network for the facades generation is the generator of the Pix2pix while the one for the cityscapes is the CRN. }
\end{figure}

\section{Conclusion}
In this paper, we propose a diversity loss objective to synthesize diverse realistic images, corresponding to a semantic layout. 
Moreover, since we link the input noise to the semantic segments in the synthesized images, we can not only synthesize diverse images, but also obtain the desired images by adjusting the noise manually. 
Experimental results show that our approach can produce a diverse set of images than the existing methods. 
Meanwhile, using our approach will not degrade the reality of the original model.

There are a few possible future directions that one can follow. 
First, note that we only use the information of semantic-level segmentation. 
There is no difficulty to generalize this idea to the one using the information of instance-level segmentation. 
Second, it seems that one can consider linking higher dimensional random noise to each semantic segment. 
But we are not sure that how one can give an explaination to the noise. 
Third, we also notice that our approach mainly induces illuminance variation or color variation. 
But the users may also require more structural variations.
This may be an interesting direction next.
Finally, one can generalize our idea to other image translation tasks without semantic layouts. 
One way is to extract high-level features of a pretrained deep neural network (for example, VGG). 
Then, one can link the random noise to the features using the same way as ours.

\appendices
\section{More Examples of the Synthesized Images}
\noindent\textbf{Facades Generation:}

The images in Figure 8, gives a case that our approach can learn some structural variations. 
\begin{figure}
\centering
\includegraphics{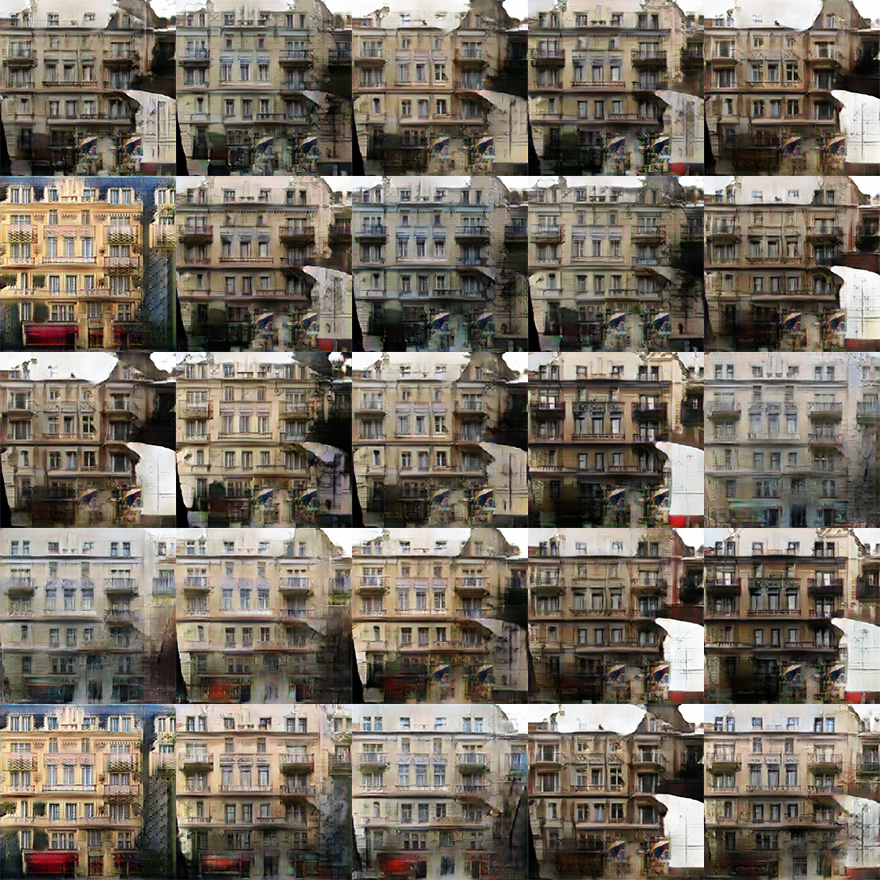}
\caption{\textbf{More examples of the generating images}: One can observe more structural variations of the facades.}
\end{figure}

The images in Figure 9, gives a case with different wall textures and window types.
\begin{figure}
\centering
\includegraphics{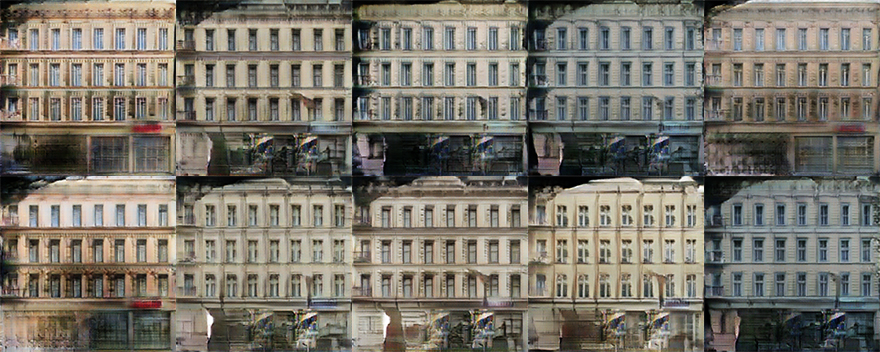}
\caption{\textbf{More examples of the generating images}: One can observe images with different wall textures and window types}
\end{figure}

The images in Figure 10, gives a somewhat bad case. Since the overall structure of this facade is too simple, the number of the generated types is correspondingly less.
\begin{figure}[!t]
\centering
\includegraphics{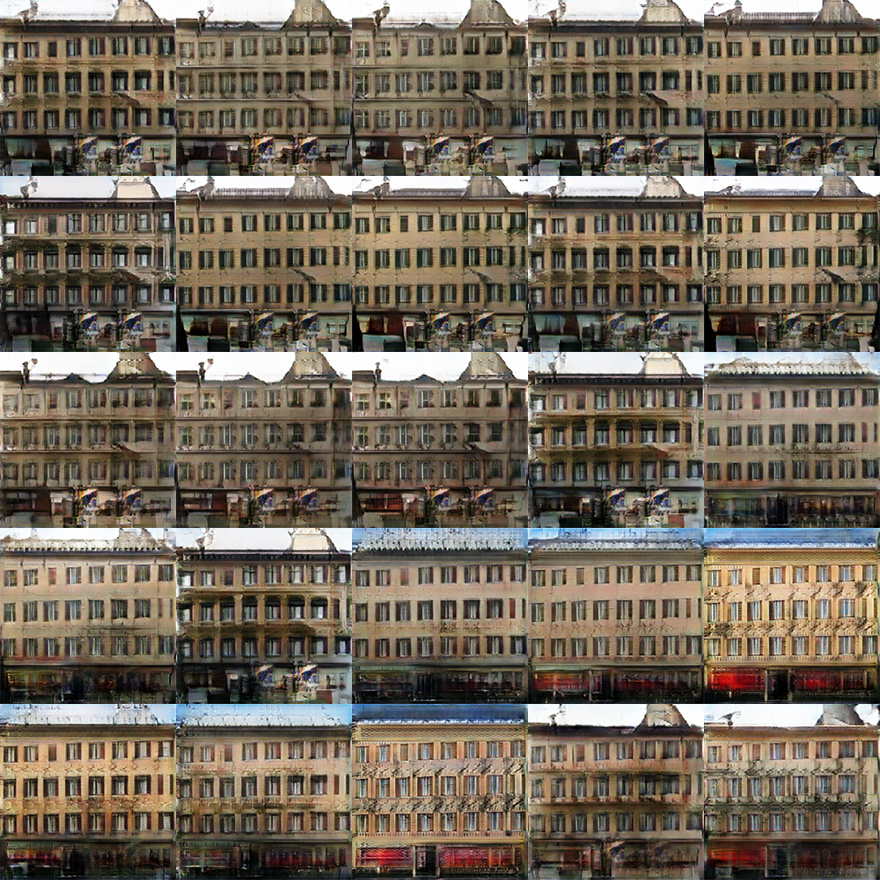}
\caption{\textbf{More examples of the generating images}: The simplicity of the semantic layout leads to a somewhat bad case.}
\end{figure}

\vspace{0.5cm}\noindent\textbf{Cityscapes Generation}:

The images in Figures 11, gives a bad case in which the cars are distorted.
\begin{figure}[h]
\centering
\includegraphics{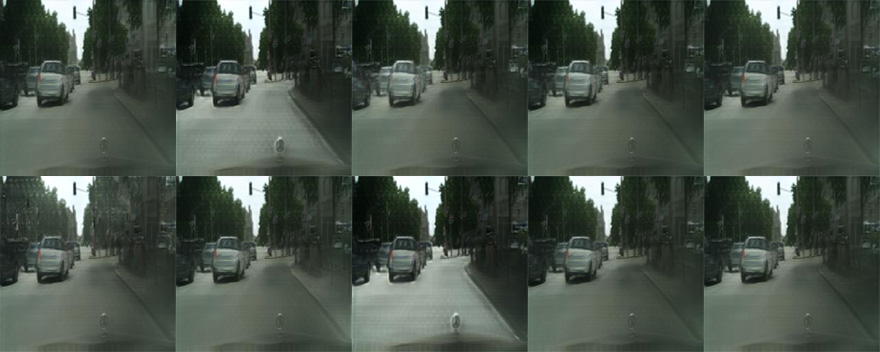}
\caption{\textbf{More examples of the generating images}: The generated cars are distorted.}
\end{figure}

\section{Algorithm for Training Networks with the Diversity Loss}

In this section, we give a brief description on the training of networks with the diversity loss, which is a little different from the typical network training process.
The main difference is the lines 6-9, where we feed forward two times(One can also concatenate the inputs as a batch of size two).

\begin{algorithm}
\renewcommand{\algorithmicrequire}{\textbf{Input:}}
\renewcommand{\algorithmicensure}{\textbf{Output:}}
\caption{Training Networks with the Diversity Loss}
\begin{algorithmic}[1]
\REQUIRE training data $\{l_k, i_k\}_{k=1}^{N}$, base network $f$, hyperparameters $\beta, \lambda_c$ for each $c\in C$, number of epochs $E$.
\ENSURE trained model $f_\Theta$.
\STATE Randomly initialize $\Theta$.
\FORALL {$epoch\leq E$}
\FORALL {$k\leq N$}
\STATE Generate random noise $noise\in[-1,1]^{|C|}$.
\STATE Use $noise$ and $l_k$ to generate noise input channel $l_k'$(See the description in Experiments).
\STATE Feed the network $f_{\Theta}$ with $(l_k, l_k')$ to get output $i_{k,1}$.
\STATE Feed the network $f_{\Theta}$ with $(l_k, 0)$ to get output $i_{k,2}$, where $0$ means a channel filling with zeros.
\STATE Compute the diversity loss $L_{Div}$ between $i_{k,1}$ and $i_{k,2}$
\STATE Update $\Theta$ with the gradient induced by $L_{f}+\beta L_{Div}$ where $L_{f}$ is the loss of the base network $f$.
\STATE k = k + 1
\ENDFOR
\STATE epoch = epoch + 1
\ENDFOR

\STATE \textbf{return} $f_\Theta$
\end{algorithmic}  
\end{algorithm}


\bibliographystyle{IEEEtran}
\bibliography{IEEEabrv,reference.bib}

\begin{IEEEbiography}[{\includegraphics[width=1in,height=1.25in,clip,keepaspectratio]{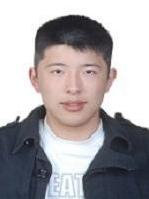}}]{Zichen Yang}
is a master candidate in the Department of Computer Science, Zhejiang University. He got his Bachelor degrees in Computer Science and Mathematics at Chongqing University. His current research interests include image processing, computer vision and artificial intelligence.
\end{IEEEbiography}

\begin{IEEEbiography}[{\includegraphics[width=1in,height=1.25in,clip,keepaspectratio]{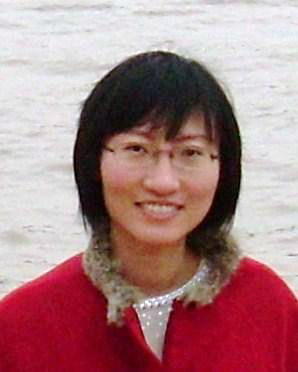}}]{Haifeng Liu} is an Associate Professor in the College of Computer Science at Zhejiang University, China. She received her Ph.D. degree in the Department of Computer Science at University of Toronto in 2009. She got her Bachelor degree in Computer Science from the Special Class for the Gifted Young at University of Science and Technology of China. Her research interests lie in the field of machine learning, pattern recognition, and web mining.
\end{IEEEbiography}

\begin{IEEEbiography}[{\includegraphics[width=1in,height=1.25in,clip,keepaspectratio]{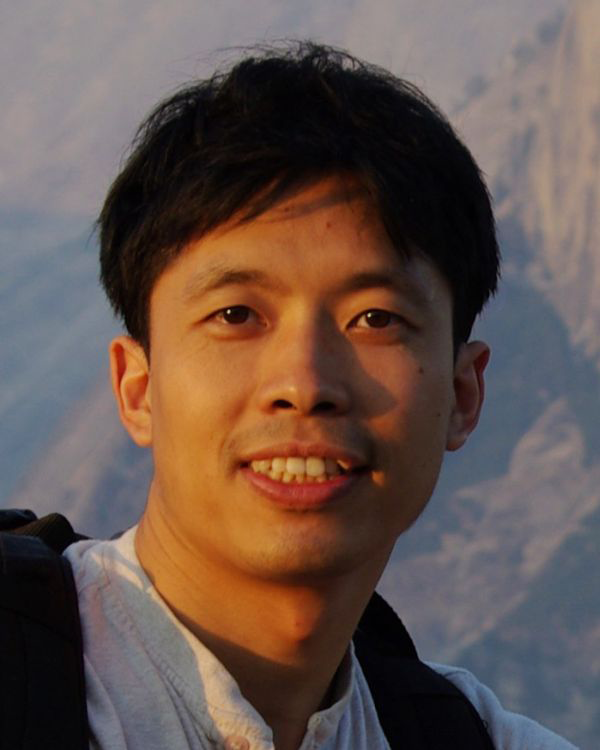}}]{Deng Cai} is a Professor in the State Key Lab of CAD\&CG, College of Computer Science at Zhejiang University, China. He received the Ph.D. degree in computer science from University of Illinois at Urbana Champaign in 2009. His research interests include machine learning, data mining and information retrieval.
\end{IEEEbiography}

\vfill


\end{document}